\documentclass[10pt,letterpaper]{article}
\usepackage[letterpaper,margin=1in]{geometry}
\usepackage{newtxtext}
\usepackage{natbib}
\setcitestyle{authoryear,round,citesep={;},aysep={,},yysep={;}}
\usepackage{amsmath,amsfonts,bm}

\def\eqref#1{equation~\ref{#1}}

\def\1{\bm{1}}

\DeclareMathAlphabet{\mathsfit}{\encodingdefault}{\sfdefault}{m}{sl}
\SetMathAlphabet{\mathsfit}{bold}{\encodingdefault}{\sfdefault}{bx}{n}

\usepackage[T1]{fontenc}
\usepackage[utf8]{inputenc}
\usepackage{microtype}

\usepackage{amssymb}
\usepackage{newtxmath}

\usepackage{graphicx}
\usepackage{booktabs}
\usepackage{hyperref}
\hypersetup{pdftitle={Close Enough is Nowhere Near: Fidelity-Consumability Decoupling in KV Caches},pdfauthor={Adam Deng, Shicheng Liu, Yada Zhu}}

\usepackage{url}
\usepackage{xcolor}

\newcommand{\lead}[1]{\textbf{#1}}

\title{Close Enough is Nowhere Near:
Fidelity-Consumability Decoupling in KV Caches}

\author{Adam Deng\\MIT ORC \and Shicheng Liu\\MIT CSAIL \and Yada Zhu\\IBM Research}

\date{}

\begin{document}

\maketitle

\begin{abstract}
Multi-agent LLM systems communicate in text, which is lossy and wasteful: the sender throws away the reading it already did. A growing literature instead transmits the KV cache, the natural carrier of that reading. We study the most appealing form of the idea: both models stay frozen and the sender’s cache is substituted for the reader’s with nothing trained. It does not work, and the obstruction belongs to the reader. A frozen transformer reads only caches close to the one it would have written itself, a region we call its tolerable set. We show that geometric similarity to that cache does not predict whether the reader can use it, and the failure is not gradual, so a more faithful substitute is not the remedy it appears to be. We explain this phenomenon with competitive attention. Experimentally, we test eight diverse substitution methods against two measured anchors: an uninformative-cache baseline and the reader’s own accuracy from having read the passage. All sit at the baseline. That turns a long tail of separate negative results into one outcome with a stated condition for overturning it. A different design does work: if the reader keeps its own cache and attends over the sender’s as extra context, adapting only its queries, it beats its own baseline, and also clears the baseline with eight senders pooled where undirected attention over the same union falls below it. We offer this as a characterization on one probe, not a leaderboard claim. Our central contribution is a redirection: pursuing faithful substitute caches is the wrong target; adding a cache to what the reader already has is the promising one.
\end{abstract}

\section{Introduction and Related Work}

Multi-agent language-model systems communicate through generated text~\citep{moa,debate,autogen,camel,metagpt,agentsurvey}, at a structural cost: the sender discards the reading it already performed by projecting it into tokens, the reader pays for that reading again, and whatever did not survive the projection into words is unrecoverable. That bottleneck motivates transmitting the internal state computed during reading instead. A growing literature pursues it through cache-to-cache messaging over a learned gated residual~\citep{c2c}, trained latent alignment of key--value memories~\citep{latentkv}, and closed-form cache maps within a model family~\citep{xkvlinear,droidspeak}. The key--value (KV) cache is the natural candidate carrier for such a channel. A transformer constructs one during prefill regardless of whether anything is shared, so the representation is already paid for. It records what the model read in digested form instead of in words, and it already resides in memory, addressable and of known shape.

This raises a question prior to the design of any particular channel: what will a frozen model accept from another model in place of its own cache? We study the most permissive version of the proposal: both models frozen, the sender's cache handed to the reader in place of its own, no access for the sender to how the reader reads, and no fine-tuning. This is the version that composes, since any two frozen agents will do, with no joint training and no representational conventions negotiated in advance. It fails, and it fails for a reason that belongs to the receiving model, not to any particular bridge. Scattered reports of such failures exist, typically attributed to the limitations of one construction. What we find instead is a property of the reader itself: a frozen transformer reads a cache correctly only if that cache lies inside a narrow neighborhood of the one its own forward pass would have produced, and it is narrow in a sense that no norm on the cache captures. We call that neighborhood the reader's \emph{tolerable set}, and we find that nothing constructed from the sender's side lands inside it. Two findings are worth stating at this level of generality, because both run against the prevailing intuition. Geometric similarity to the reader's own cache does not predict whether the reader can use it: we exhibit pairs ordered one way by similarity and the opposite way by transfer. The failure is also not a matter of accumulation over a deep stack: corrupting a single layer is already as damaging as corrupting all twenty-eight.

The mechanism is located at the point where a cache enters the reader's computation. That point is the attention softmax, and it is competitive: keys are scored against each other, not against a threshold, so the output is determined by which tokens prevail. A displacement too small to register as geometric distance suffices to change which tokens prevail, after which the reader attends to the wrong part of its context. Section~\ref{sec:mechanism} measures this directly, and also rules out the alternative account on which off-manifold caches are catastrophically amplified. This diagnosis suggests a different design. Rather than replacing the reader's cache, retain it and expose the sender's cache as additional attendable context, with adaptation confined to the reader's query side. The reader is then never asked to consume a foreign object in place of its own, and the tolerable set does not bind. We show that this substantially improves transfer, can match the reader's native-context performance, and continues to work when several senders are pooled.

\lead{One distinction organizes the prior work.} Results that look inconsistent in isolation reconcile once three cases are separated: whether a shared cache must \emph{replace} the reader's own (substitution), is merely \emph{added} to what the reader attends over (augmentation), or comes from a map \emph{fitted} against the reader's own caches. Sections~\ref{sec:framework} and~\ref{sec:consequences} state these as claims (C1)--(C3). The internal-channel proposals above divide along exactly this axis, and the strongest of them supports rather than contradicts our account: the cache-to-cache ablation that \emph{projects} the sender's cache into the receiver's slots collapses, while the configuration keeping the receiver's cache and adding to it is the one that works. It also explains why text worked. Text re-enters through the receiver's own encoder before any attention occurs, and a borrowed cache is afforded no such entry; retrieval-style cross-attention~\citep{retro} and fusion of model \emph{outputs}~\citep{moa,debate} are augmentation on the same reading, and the division predicts that they work, as they do.

Two adjacent literatures fall on the near side of the distinction. Same-model compression shrinks a cache by evicting low-attention tokens~\citep{h2o,scissorhands}, factorizing across depth~\citep{minicache}, or quantizing entries~\citep{kvquant}. These methods succeed, and our account is consistent with that because producer and reader are the same model there, so every modification perturbs a cache the reader itself wrote and the question is asked from inside the tolerable set rather than from outside it (Section~\ref{sec:mechanism}). Work reusing caches across layers or models~\citep{minicache,droidspeak} or merging weights to pool capabilities~\citep{fusellm} reports mixed outcomes the distinction predicts: where reuse is substitution and the bridge is built without reference to the reader's own cache on the target context, we find it held to the accuracy the reader already reaches without any shared cache, and the binding constraint belongs to the reader rather than the bridge (Section~\ref{sec:bank}). A separate line measures correspondence between representations and exploits it~\citep{cka,relreps,stitching}, acknowledging but stating weakly that similarity is not interchangeability. Our results sharpen that caveat and locate its reason, for attention, in the gap between the Euclidean alignment such indices measure and the competitive readout that consumes a cache (Sections~\ref{sec:cosine} and~\ref{sec:mechanism}).

\lead{Fitted bridges, and our claim's boundary.}
\label{sec:related-transfer}
A bridge \emph{fitted} against the reader's own caches, regressed onto the very quantity it must reproduce, is different from the sender-side constructions we test. Our negative claim is scoped away from it: we run no fitted bridge, and state that as a non-claim. Published evidence makes this a real boundary. \citet{xkvlinear} fit per-head closed-form ridge maps between members of one model family, freeze both models, and have the receiver decode from the mapped cache \emph{in place of} its own, retaining most of the original quality; \citet{latentkv} train a translator constructing the receiver's prefix cache from the sender's, with large gains over an untrained identity map. Both satisfy the qualification our claim carries: the map is fitted against the reader's forward pass on the transferred context rather than built from the sender's side alone, which establishes that the qualification is not idle; it is the whole content of the claim. Both settings evaluate under \emph{redundancy} instead of information asymmetry, the receiver being scored on continuations of a context it has seen or could recompute; our probe is built to remove that redundancy, and we claim only that a fitted within-family map has, to our knowledge, not been tested there (Appendix~\ref{app:fitted}).

\section{The Object of Study}
\label{sec:framework}

The object of study is not a cache, but what a particular frozen model does with one: supply it with key and value tensors, let attention consume them in place of recomputing them, and read the output distribution. Everything downstream is fixed, the weights being fixed. The consequential point, and the origin of every difficulty that follows, is that this mapping belongs to the reader. One cache supplied to two models passes through two different functions, and a value computed by one carries no particular meaning for the other.

One cache is distinguished: $C^{*}$, the cache the reader itself produces on the true context, calibrated by construction because the weights that wrote it are the weights now reading it. How far from $C^{*}$ may a cache lie and still be read correctly? The caches so satisfying are the reader's \emph{tolerable set}, and cache sharing is the problem of placing a candidate inside it. That set has positive radius and is strongly anisotropic (Section~\ref{sec:mechanism}), so claims of narrowness below concern reachability by the methods we test rather than a singleton. Geometric closeness to $C^{*}$ admits several standard measures, cosine similarity, reconstruction error, and attention-logit correlation, and we call any of them the candidate's \emph{fidelity}. Whether the reader can consume the cache is a distinct predicate, \emph{consumability}; the two come apart (Section~\ref{sec:cosine}).

Throughout, \emph{transfer} names the dependent variable: the fraction of probe items on which the reader's greedy continuation contains the answer string (Appendix~\ref{app:probe}). \emph{On-manifold} names an object a real forward pass of this reader could have produced, which in practice means a key it did produce on some context. A construction landing in the column space of $W_{K}$ satisfies a weaker condition, named explicitly where it arises and never called manifold membership.

Two anchors, measured within each run and not assumed, frame every number. The \emph{floor}, $0.180$, is the accuracy from the reader's own cache with the answer masked out, the rate at which the question is answerable from the rest of the passage. The \emph{native-cache reference}, $0.665$, is the accuracy from its own cache of the unmasked context. Neither is a formal bound, so values outside $[0.180, 0.665]$ are possible and are flagged where they occur.

Our position: \emph{(C1) Sender-side substitution is floor-bounded.} A method that constructs a cache and installs it in the reader's slots in place of $C^{*}$ stays near the floor unless its output lands inside the tolerable set, and raising fidelity does not help. The scope is methods built \emph{from the sender's side alone}, without fitting the producing map against the reader's own forward pass on the context being transferred; every arm of our sweep observes that discipline. \emph{(C2) Augmentation is not bound the same way.} A method that leaves $C^{*}$ in place, appends the sender's cache as further attendable context, and adapts only the query side never asks the reader to consume a foreign cache in place of its own. It falls outside the regime the tolerable set constrains and can approach the native-cache reference. Both claims admit refutation, and Section~\ref{sec:falsify} states the refuting observation for each alongside the experiments most likely to produce it.

\section{Three Measurements of the Tolerable Set}
\label{sec:geometry}

Every measurement in this section uses one reader and one probe. The reader is Qwen2.5-1.5B~\citep{qwen}, 28 layers with grouped-query attention at group size $G{=}6$, frozen throughout. The probe is split-context factual recall built from SQuAD: a sender has read a passage containing the answer, the reader receives the same passage with the answer masked together with the question, and the reader is supplied a cache derived from the sender and scored on whether it produces the answer. Decoding is greedy over eight tokens at $n{=}200$ items unless stated otherwise, and the full specification with both anchors is Appendix~\ref{app:probe}. The three measurements ask how faithful a cache must be, how much of it must be correct, and whether selecting a real cache instead of synthesizing one evades the constraint.

\subsection{Cosine Similarity Does Not Predict Consumability}
\label{sec:cosine}

The most direct formulation displaces the reader's own cache by a calibrated distance and finds where the reader fails. The content is the reader's own masked cache, which carries no answer, so tolerance of the geometry alone would register at the floor. Tolerance turns out to be minimal. At cosine $0.9999$ accuracy sits at the floor, $0.180$; at $0.999$ it has fallen to $0.120$; at $0.995$ it reaches $0.005$ and stays there for every smaller cosine tested. That last value warrants attention, being not the floor but far below it: a cache $99.5\%$ aligned with the reader's own leaves it performing worse than one carrying no information at all, having destroyed its use of the context still available to it.

What makes this a claim about the instrument rather than about noise is the comparison against a \emph{genuine} cache, the one this same frozen model produces on the unmasked passage. At full strength it sits at achieved cosine $0.9949$, \emph{further} from the reader's own cache than the noise condition scoring $0.005$, and it transfers at $0.445$ (Figure~\ref{fig:tolerance}(a), Table~\ref{tab:cosine}). Cosine does not predict consumability weakly; it orders these two conditions backwards. The governing property is closer to provenance than to distance: what matters is whether a real forward pass of a model with this geometry produced the object, and no norm-based metric detects that.

The methodological consequence bears on a substantial literature, where reconstruction error and cosine are standard instruments for evaluating cache reconstruction, compression, and sharing. On this evidence they can be driven arbitrarily close to their optima while the capability they proxy is absent, and they can rank a working method below a broken one. A transfer result reported in cosine or mean-squared error alone is uninformative as to whether the intended reader can consume it. Such metrics remain useful as diagnostics, and we use them as such, but behavioral evaluation on the reader that will consume the cache is what settles the question. We show this for one reader and offer it as a recommendation, not a proven universal.

\lead{Sparsity of the perturbation.} Production methods reproduce most features approximately and a few badly, and attention might route around a few corrupted coordinates. Neither form of the objection survives: the layer experiment below is the structural analogue, holding 27 of 28 layers bit-exact, and pairing each sparse condition with a dense perturbation at matched achieved cosine, the two agree throughout (Appendix~\ref{app:sparse}). The result above needs no narrowing to dense perturbations.

\subsection{A Single Corrupted Layer of Twenty-Eight}
\label{sec:onelayer}

The second measurement concerns how much of a cache must be correct. We corrupt the first $\ell$ layers and take layers $\ell{+}1$ through $28$ unchanged from the reader's own forward pass, sweeping $\ell$ from $0$ to $28$ under two corruptions at matched mean key cosine $0.822$: isotropic noise, which no forward pass would produce, and replacement by the nearest genuine key from a bank the reader generated on other passages. The off-manifold curve of Figure~\ref{fig:mechanism}(a) carries the result. One corrupted layer takes the reader from $0.665$ to $0.030$, while the remaining 27 layers stay bit-identical to what it wrote, not approximations of it, and that counts for nothing. The curve is flat from $\ell{=}1$ to $\ell{=}28$: the damage is done at once and does not accumulate with depth.

What this excludes should be stated directly, the compounding account being a natural reading of the failure and a widely held one. Were the mechanism a small per-layer error with survival factor $\pi < 1$ raised to the depth, one corrupted layer would cost about $\pi$ and the curve would bend downward as $\ell$ grew. That predicts a gradual slope; the measurement is a step function, so the account is discarded rather than qualified. The on-manifold row supplies the control, behaving at matched cosine exactly as compounding describes: one genuine substituted layer is survivable and slightly beneficial at $0.695$, three layers reach $0.745$, decay sets in around $\ell{=}8$, and zero arrives only once all twenty-eight are replaced. Depth-wise accumulation is therefore real and observable in this sweep, and it is not what destroys a borrowed cache. Two corruptions identical in fidelity, differing only in whether a real forward pass could have produced the injected object, diverge from fatal within one layer to surviving eight. The distinguishing property is neither geometric distance nor depth.

To check whether this reflects the reader's particular depth, we ran the substitution test across seven models spanning 6 to 36 layers: Pythia-70m/410m/1.4b, Qwen2.5-0.5B/1.5B/3B, and GPT-2. Full substitution by nearest genuine keys transfers at or below $0.008$ in every case, at $n{=}120$ against each model's own anchors. The shallowest reader bounds the claim without confirming it, its own native-cache reference leaving almost no headroom, so the informative cases run from twelve to thirty-six layers (Appendix~\ref{app:layers}). A shallower reader is not more permissive, which is the expected outcome if the failure occurs at a consumption point present in every layer.

\subsection{Selection from Six Thousand Genuine Keys}
\label{sec:bank}

Both preceding measurements construct a cache, which invites the response that construction is the difficulty: any synthesized object will be subtly wrong in consequential ways, so the operation should instead be \emph{selection}. Choosing from a library of real caches guarantees the result lies on the reader's own manifold, the reader having produced every candidate. The library holds six thousand genuine keys per layer, harvested from this reader's forward passes over other passages; for each slot we retrieve the $k$-th nearest and interpolate toward it with weight $\alpha$, so that at $\alpha{=}1$ every key attended over is a real key the reader produced on a different passage.

Selection provides no relief (Figure~\ref{fig:tolerance}(b); construction and full $k \times \alpha$ grid in Appendix~\ref{app:bank}). The nearest genuine key, the closest real object among six thousand, transfers $0.000$, and by $\alpha{=}0.5$ already $0.005$. Sweeping the rank out to $k{=}16$ leaves the pattern unchanged, so no graded reward for proximity exists among on-manifold candidates. This is our sharpest statement of how narrow the set is, and it should be distinguished from a claim with which it is often conflated. Manifold membership, in the sense fixed in Section~\ref{sec:framework}, is not the operative property: every candidate here is a key this reader actually produced, and none is consumable. The reader does not need a point of the right kind, and a nearby point will not do either, the nearest of 6,000 genuine keys being already too far. Membership is necessary and far from sufficient. A broader version of the same idea reinforces the reading. Nine variants crossing three codebook constructions, the real-key bank, a learned vector-quantized codebook, and a PCA lattice, with three snapping rules, yield a best cell at the highest cosine fidelity we achieved by pure selection, $0.880$, with attention-logit correlation $0.928$. Yet this best cell transfers $0.000$, despite having a higher cosine similarity. Among selection methods fidelity and consumability come apart and then run against each other.

\begin{figure}[th]
  \centering
  \includegraphics[width=1\linewidth]{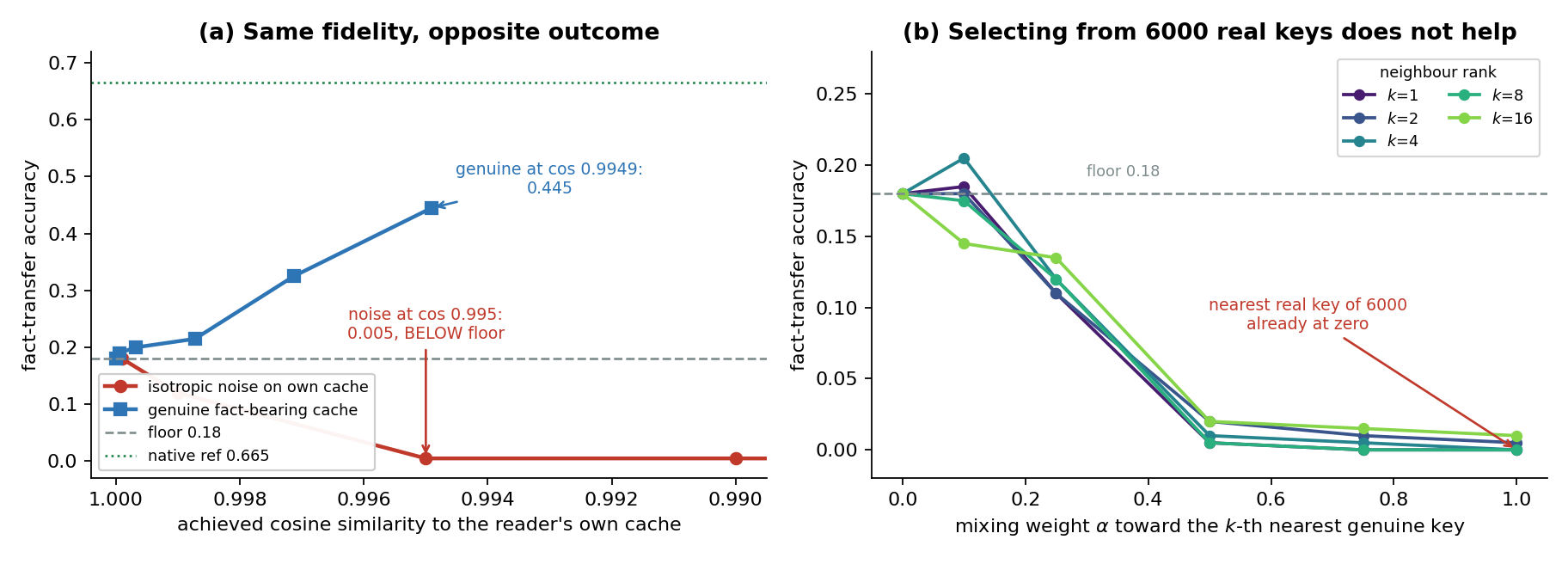}
  \vspace{-10px}
  \caption{Cosine similarity does not predict consumability. (a) Isotropic noise on the reader's
  own cache falls below the floor by cosine $0.995$, while the genuine cache the same frozen model produces on the unmasked passage transfers at $0.445$ from a \emph{lower} cosine of $0.9949$, an
  $89\times$ gap at matched fidelity, ordered inversely. (b) Retrieval from a bank of 6000 genuine
  keys: tolerance is flat in neighbor rank $k$ and collapses in the mixing weight. The nearest real
  key among six thousand fails.}
  \label{fig:tolerance}
\end{figure}

\section{Why Substitution Cannot Reach the Tolerable Set}
\label{sec:mechanism}

Three measurements converge on one property. This section accounts for it, locating the failure not in the cache but in the operation that consumes it. We begin with the explanation one reaches for first.

\subsection{Not Amplification}

The intuitive account posits a cliff: off-manifold displacements are amplified catastrophically en route to the output while on-manifold ones are not. It is testable, and it is wrong. Measured at both stages through which an injected key acts, the off-/on-manifold readout gain is an $O(1)$ ratio, moderately more consequential and not catastrophically so, with no layer exhibiting a blowup and the logit stage attenuating instead of amplifying (Figure~\ref{fig:mechanism}(b); measurement in Appendix~\ref{app:cond}). The sensitivity that does exist is architectural and analytically pinned, its governing constant set by the grouped-query attention~\citep{gqa} group size: several query heads read each key slot, so a displacement there is read out once per head. That factor is real, and far too small to account for the collapse. Whatever destroys a borrowed cache does not do so by enlarging the displacement.

\subsection{The Softmax Consumption Point}

The constraint arises where the cache enters the reader's computation. A cache couples to behavior through one operation,
\begin{equation}
  \mathrm{Attn}(q, K, V) = \mathrm{softmax}\!\left(\frac{q K^{\top}}{\sqrt{d}}\right) V,
\end{equation}
and that operation is \emph{competitive}: keys are scored against one another, not against a fixed threshold, so what the reader reads is decided by which keys win and by what \emph{margin}, never by how close the cache is in any norm. A displacement far too small to register geometrically can flip a decisive contest, while a larger one along an uncontested direction passes unnoticed. This is the explanation, and the rest follows from it. The contests are also numerous, one per query, per head, and per layer, so a modest context stages tens of thousands at once. Tolerance means surviving every one.

Three measurements are consistent with the account and hard to reconcile with its alternatives. At matched cosine, a genuine cache is consumable and a noise-calibrated one is not, so what matters is whether the displacement respects the margin structure the reader's own forward passes produce, and not how far it moves (Section~\ref{sec:cosine}). Manifold membership fares no better: it constrains where a key lies, not which contest it wins, and a genuine key is genuine for some other context (Section~\ref{sec:bank}). Nor does anything compound over depth, one layer's attention already being a competition whose outcome the intact layers downstream faithfully process (Section~\ref{sec:onelayer}).

\lead{The tolerance is anisotropic, not uniformly small.} This formulation best fits the evidence, and it reconciles our negative result with a literature of positive ones. Tolerance is wide along directions the reader's attention does not resolve and thin along those it does. A norm on the cache averages over both and reports neither. Same-model compression, whether by eviction, factorization, or low-bit quantization, operates on a cache the reader itself wrote, and in succeeding demonstrates that the wide directions really are wide. Cross-model production has no reason to respect the orientation, and lands in the thin ones. The open question is therefore not how to make a produced cache closer, but how to confine its residual error to directions the reader cannot resolve, and nothing in our sweep suggests the sender's side has the information required.

Fragility is thus a property of the readout, not of the representation or of the stack. It is the generic consequence of handing a candidate object to a normalized competitive operator. A KV cache is reader-specific, portable only through a mechanism with access to how the reader will consume it, a qualification Section~\ref{sec:related-transfer} shows is doing real work. The account predicts one asymmetry, and it holds: keys set the competition while values are averaged under weights already fixed, so key damage costs more. At matched sparsity, destroying thirteen of two hundred fifty-six value coordinates is free while the same damage to keys is fatal, and cosine orders the two backwards (Appendix~\ref{app:sparse}). The account also supplies a design lever, developed next. If a foreign cache cannot survive consumption \emph{in place of} the reader's own, it should not be put there.

\begin{figure}[th]
  \centering
  \includegraphics[width=1\linewidth]{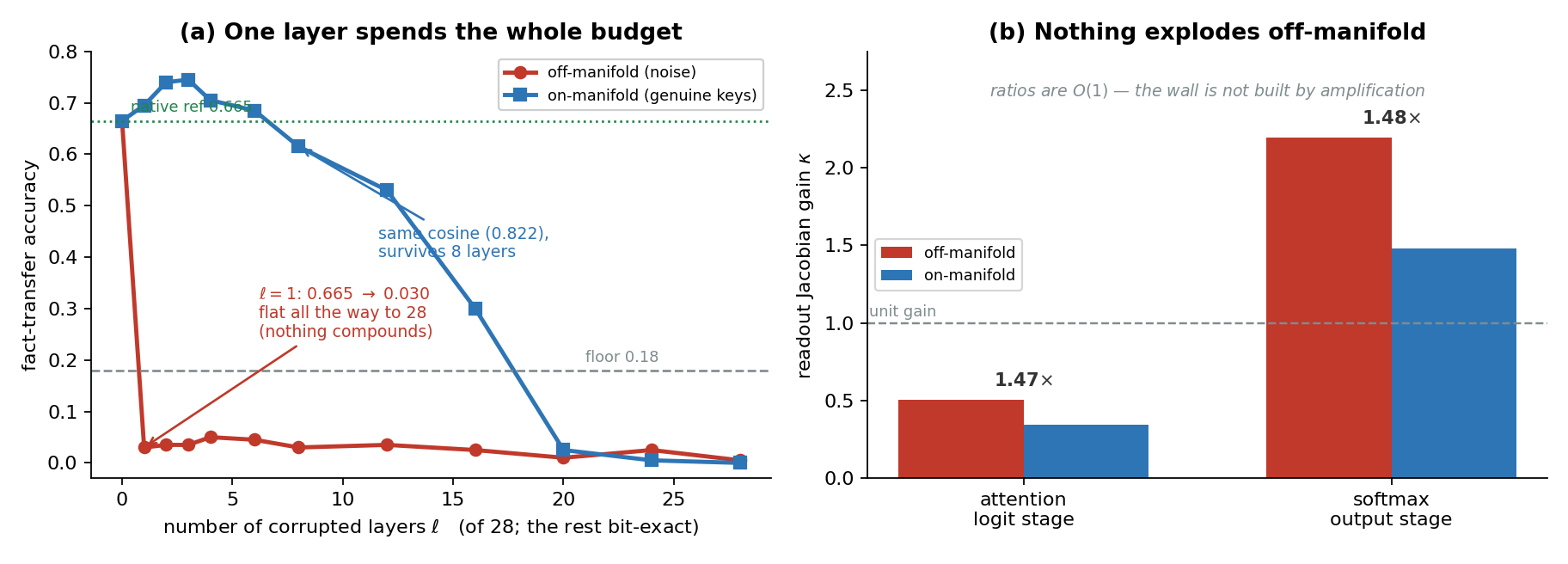}
  \vspace{-10px}
  \caption{Two explanations that do not survive measurement. (a) Corrupting the first $\ell$ of 28
  layers: off-manifold damage exhausts its budget at $\ell{=}1$ and is flat thereafter, so nothing
  accumulates over depth, while on-manifold substitution at matched cosine survives eight layers and
  decays gradually. (b) The off-/on-manifold readout-gain ratio is $1.47\times$ and the logit-stage gain
  falls below one, so the effect is not constructed from amplification either.}
  \label{fig:mechanism}
\end{figure}

\section{Consequences: Two Regimes Run and a Third Surveyed}
\label{sec:consequences}

The characterization predicts a division, not a gradient, and the sweep below tests it by varying \emph{how the shared cache is produced and where it is placed}, holding the reader, the probe, and the anchors fixed. Under \emph{substitution} the produced cache occupies the reader's own key/value slots. Under \emph{augmentation} the reader keeps its own and attends over the sender's as additional context. Substitution must land inside the tolerable set and should therefore be floor-bounded however it is built from the sender's side alone. Augmentation never faces that requirement, and since the reader ends with strictly more to attend over, it should be able to reach the native-cache reference.

A third regime lies outside the sweep: a \emph{fitted bridge}, whose producing map is regressed against the reader's own caches, not built from the sender's side. We state it as a non-claim, that \emph{(C3) fitted bridges are not floor-bounded, and we do not run one}. Section~\ref{sec:related-transfer} reviews published evidence for it.

\subsection{Substitution, Eight Ways}
\label{sec:sweep}

The methods below function as instruments and not as proposals, and we intend no leaderboard claim. The evidence lies in their agreement: unrelated constructions, evaluated on one probe against one pair of anchors, return the same reading of the same property of the reader.

We varied the production method across four orders of magnitude of capacity and four families of objective, from a freeform regression onto the reader's cache to objectives driving produced keys onto its key manifold, into the column space of $W_{K}$, or against a worst-case per-layer hinge, and from regression to selection. Table~\ref{tab:sweep} lists them. Two rows carry more than their own value. The highest-capacity method discards the low-rank bottleneck for a per-layer, full-width map and reaches $0.200$, the strongest substitution result in the sweep, which places it at the floor. The oracle supplies the nearest genuine key from the bank, provably on the reader's own manifold, and transfers $0.000$: genuine provenance confers no consumability in the slots.

\begin{table}[th]
\centering
\label{tab:sweep}
\small
\begin{tabular}{llcl}
\toprule
Regime & Production method & Transfer & Note \\
\midrule
Substitution & freeform regression                & $0.000$  & trained, low-rank \\
Substitution & manifold-loss training             & $0.010$  & trained toward the key manifold \\
Substitution & by-construction (column space)     & $0.010$  & in the column space of $W_{K}$ \\
Substitution & row-space target                   & $0.085$ & trained; see note on spread \\
Substitution & worst-case hinge                   & $0.080$  & trained, per-layer hinge \\
Substitution & high-capacity per-layer MLP        & $0.200$  & best substitution; at the floor \\
Substitution & codebook selection                 & $0.000$  & best selection fidelity (cos $0.880$) \\
Substitution & oracle nearest genuine key         & $0.000$  & real key, 6000-key bank \\
\midrule
Augmentation & cross-attention, no adaptation     & $0.555$ & zero trained parameters \\
Augmentation & cross-attention + query-side LoRA  & $\mathbf{0.695}$ & \emph{at reference} $0.665$ ($z{=}0.6$) \\
\bottomrule
\end{tabular}
\vspace{-5px}
\caption{The sweep. Substitution methods built from the sender's side remain at or below the floor regardless of capacity, objective, or whether the produced cache is constrained to lie in the reader's key column space. Both augmentation methods clear it, and the adapted variant reaches the native-cache reference. Floor $0.180$, native-cache reference $0.665$, $n{=}200$ throughout.}
\end{table}

The substitution block is the central negative result, and its force derives from the flatness rather than from any row. Methods sharing nothing in their construction, trained and untrained, bottlenecked and full-width, regression and retrieval, on-manifold by proof and off-manifold by accident, all land at or below the floor. Because every row is measured on one probe against one pair of anchors, this is not a collection of independent engineering shortfalls that happened to coincide; a shared bound reached by unrelated methods indicates a property of the reader. It also answers one reading of the fitted-bridge results of Section~\ref{sec:related-transfer}, that they succeed by having more capacity to spend: our full-width arm has the capacity and still lands at the floor, so what separates a fitted bridge from these arms is what it is fitted against. A further diagnostic sharpens the method-independence. For each produced cache we count how many of the reader's 28 layers receive an attention-logit correlation above $0.99$ against the genuine cache, which is a count of how many layers are read approximately as intended. The identity injection scores 28 of 28. No substitution method we constructed reaches double digits, and within the trained-reconstruction family the count is pinned more tightly still: five variants spanning the by-construction, row-space, worst-case, and high-capacity objectives all return exactly 3, the full-width map buying no additional layer over the low-rank one (Appendix~\ref{app:sweep}). Capacity moves the fidelity metrics and leaves this count unchanged.

\subsection{Augmentation Clears the Floor and Reaches the Native-Cache Reference}

Changing only the placement, the reader retains its own cache $C^{*}$ and attends over the concatenation $[C^{*}; C_{\mathrm{sender}}]$, with the sender's cache passed through verbatim. There is no reconstruction, no projection, and no bridge trained toward resemblance. With no trained parameters this transfers at $0.555$, three times the floor and beyond the reach of the entire substitution family. Adapting the reader's query-side projections with LoRA~\citep{lora} at rank 16 raises it to $0.695$, at the native-cache reference of $0.665$. At $n{=}200$ that crossing is six items ($z=0.64$), so we report it as reaching the reference, not as beating it. The comparison that carries weight is against the floor, where the zero-parameter arm alone stands at $z=7.8$.

The gap is not a single-variable contrast, and we do not present it as one. Substitution removes $C^{*}$ while augmentation keeps it and adds to it, so the augmented reader has strictly more to attend over, and part of the gap follows from that and not from placement alone. The closest matched contrast we have is the genuine-cache arm of Appendix~\ref{app:cosine}: an answer-bearing cache this same model wrote on the unmasked passage reaches $0.445$ substituted into the slots, against $0.555$ for the same kind of object supplied as attendable context. Both arms carry the answer, so the comparison holds content roughly fixed and varies placement, and the placement is worth more than the fidelity. The oracle and augmentation arms are not the same object in two locations, the bank key carrying no answer, so we rest nothing on that pair beyond the provenance point.

We note two qualifications. (1) The strongest arm carries the least clean negative control, and control-subtracted the keys-and-values variant slightly exceeds it, so the ordering among adapted variants should not be relied upon (Appendix~\ref{app:sweep}); what survives the controls is the regime gap, an order of magnitude larger. (2) Sender and reader are the same frozen checkpoint here, reading different contexts (Appendix~\ref{app:probe}), so this is placement within one model and not communication between two.

\subsection{Scaling the Sender Pool}

A division holding for a single sender is of limited use to a multi-agent system, so we scaled the pool: the reader attends over the union of $N$ senders' caches, exactly one of which has read the passage containing the answer (Appendix~\ref{app:pool}). The bridge is trained at $N{=}4$ alone, so $N{=}1$ and $N{=}2$ are transfers out of that condition and $N{=}8$ is extrapolation beyond it.

\begin{figure}[ht]
  \centering
  \includegraphics[width=0.85\linewidth]{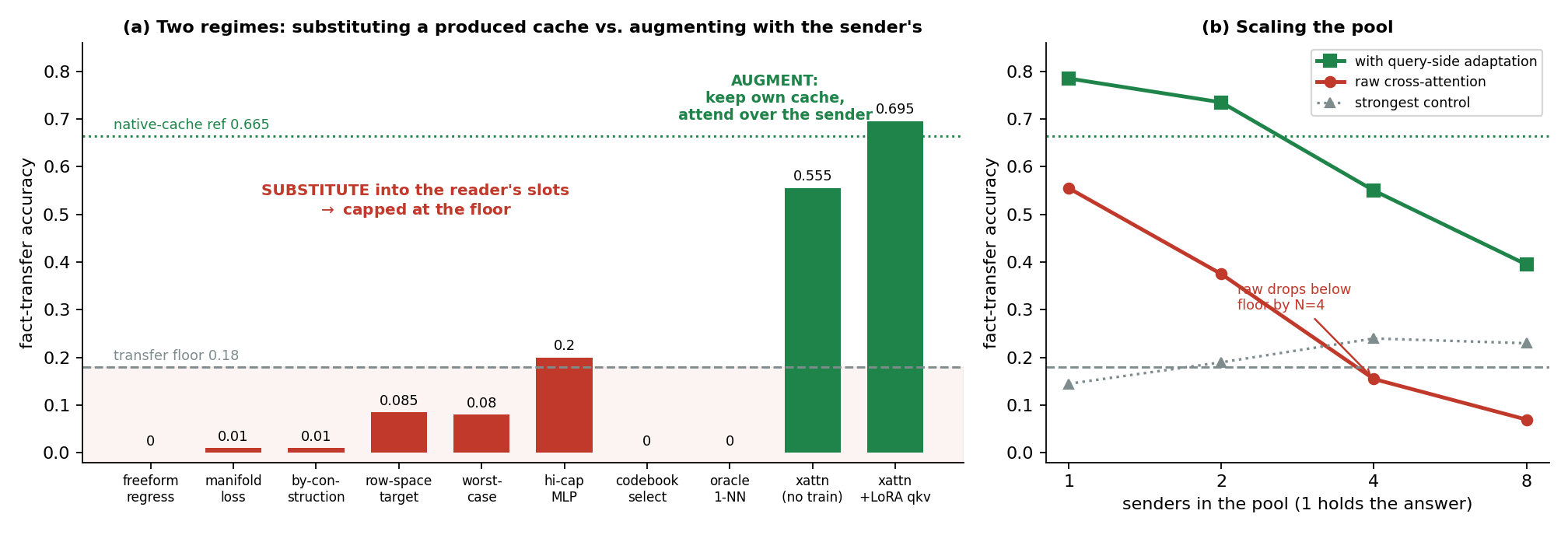}
  \vspace{-10px}
  \caption{The sweep over how the shared cache is produced and where it is placed. (a) Every
  substitution method sits at or below the floor (0.180), including an oracle supplied a real key from
  a bank of 6000. Augmentation escapes, reaching the native-cache reference (0.665). (b) Scaling the
  sender pool. Details in Table~\ref{tab:sweep} and Appendix~\ref{tab:pool}.}
  \label{fig:landscape}
\end{figure}

Figure~\ref{fig:landscape}(b) shows the two arms diverging as the pool grows, and the full table is Appendix~\ref{app:pool}. Undirected attention over a union degrades rapidly and falls below the floor by four senders, the distractors dominating. The adapted bridge decays gradually instead, from $0.785$ at $N{=}1$ to $0.395$ at $N{=}8$, still clearing the floor at eight senders with a control-subtracted lift of $+0.165$. We report control-subtracted values because the controls climb with $N$, a larger pool admitting more scoring by chance, and the runs at $N{=}4$ and $N{=}8$ flagged their own controls as elevated. What the two arms establish is that query-side adaptation is what keeps a pooled cache useful as $N$ grows. We did not measure where the adapted bridge places its attention, so we do not claim selectivity as the route.

\subsection{What the Sweep Licenses}

The division between the two regimes dominates every other source of variation: where the cache is placed decides the outcome, and how well it was built does not. Capacity, objective, and even a guarantee that the produced cache lies on the reader's own manifold leave substitution where it started, and the fine structure inside each regime is immaterial against the distance between them. We do not claim the regimes differ in kind as a matter of principle, for the reason given above.

The design rule follows from the mechanism and not from a benchmark result. A cache should not be reconstructed, selected, or synthesized to stand in for another model's own. It should be supplied as context and permitted to re-enter through the reader's own query pathway.

\section{Limitations and Falsification}
\label{sec:falsify}

\lead{Scope.} All numbers we report derive from one reader and one instrument: a split-context factual-recall probe. The anchors and sensitivity constants are that reader's properties, and the anchors are measured baselines, not proven bounds. The probe is also a single \emph{kind} of task, and its character may matter as much as its identity: short factual recall turns on whether one span wins attention, plausibly the condition most sensitive to the margin reordering we identify as the mechanism. Tasks spreading their dependence across many tokens may be more forgiving. Our claims are scoped accordingly, to a frozen reader, sender-side construction, and short factual recall under information asymmetry, with the depth ladder the only axis on which the reader varies. Augmentation results are same-architecture. Across families they fail at dimension matching, so we make no cross-family claim.

\lead{Falsifiers.} (C1) fails if some substitution method transfers well above the floor while its output demonstrably lies outside the tolerable set, severing consumability from membership. (C2) fails if augmentation with query-side adaptation cannot clear the floor. (C3) marks a boundary, not a prediction: a bridge fitted by regression against the reader's own caches is not claimed to be floor-bounded, and we do not run one. Because we attribute the wall to the softmax and not to depth, raising readout sharpness should contract the tolerable set while adding layers at fixed sharpness should not, the two being ordinarily confounded and worth varying independently. A reader consuming its cache somewhere other than a softmax should also be more permissive, and tolerance should widen as a task's answer depends less sharply on individual attention margins. The sparsity of the perturbation (Section~\ref{sec:cosine}) has been tested at the feature level and does not favor the objection (Appendix~\ref{app:sparse}); the sharpest outstanding test is the carrier comparison noted below.

\lead{A prediction the account makes.} If the obstruction is a competitive normalized readout, a carrier consumed \emph{before} any softmax should prove markedly more permissive. Residual-stream activations are such a carrier, read through normalization, a linear map, and an additive residual. Our preliminary work with them is consistent with markedly greater tolerance, by roughly the margin the mechanism would predict (numbers and caveats in Appendix~\ref{app:acts}). We report it as a direction, not a result, because those runs carry their own anchors and compare across sharing protocols instead of holding displacement matched. The controlled version is the sharpest way to falsify the mechanism: if a pre-softmax carrier proved no more permissive than keys, the account would be wrong.

\lead{Conclusion.} A frozen transformer reads a cache handed to it in place of its own only when that cache is very nearly the one it would have written itself. The tolerable set has positive radius, but that radius is not what any fidelity metric reports, and it is small along exactly the directions attention resolves. What this excludes is replacement. What it leaves is augmentation, which reframes the motivating objective. Multi-agent systems need one model to attend over another's state, not to overwrite its own with it. Internal communication between frozen models is not blocked in principle. It is blocked when posed as replacement under a sender-side construction discipline.

\section*{Use of AI Assistance}

In this work, we used generative AI tools to implement/code some methods, to design and give feedback on experiments, to propose and refine hypotheses, to support analysis of results, and to help interpret results. We also used them to draft and edit parts of this paper, to create the figures from stored result records, to summarize and search existing literature, and to format references. We have not used generative AI tools to generate synthetic datasets, to develop theoretical models, or to formulate or prove mathematical claims. The one analytic expression we report, the grouped-query
readout constant of Appendix~\ref{app:cond}, was derived by hand and checked against measurement.

We have reviewed all AI-assisted work. Every quantitative claim was traced to the stored result record for the run that produced it, and claims that could not be so traced were removed. We read all AI-generated experiment code and checked its controls and parity checks against the manipulations they were meant to implement. We verified AI-drafted prose against the underlying results, and checked that every cross-reference resolves to a section containing the evidence cited. We take responsibility for the final content of this work, including text, claims, and artifacts produced with the aid of generative AI.

\bibliographystyle{iclr2027_conference}
\bibliography{references}

\appendix

\section{Probe and Anchors}
\label{app:probe}
The probe is split-context factual recall built from SQuAD. For each item we take a passage containing a short answer (at most four words, appearing verbatim in the passage), and build two versions: the full passage, and the same passage with the answer string replaced by \texttt{\_\_\_\_}. The prompt is the passage, then \texttt{Question:}, then the question, then \texttt{Answer:}. A sender reads the full version; the reader gets the masked version. We reserve the final token of the prompt so that decoding begins from a cache of the preceding tokens, generate eight tokens greedily, and score a hit if the normalized answer string appears in the normalized generation. Contexts are capped at 256 tokens and $n{=}200$ evaluation items are drawn from the validation split.

The reader throughout is Qwen2.5-1.5B: 28 layers, grouped-query attention with 12 query heads over 2 key/value heads (group size 6), head dimension 128, hidden width 1536, held frozen. In the augmentation arms the sender is a second instance of the same frozen checkpoint reading a different context, so those results are within-checkpoint and we claim nothing broader; the cross-family arm replaces it with GPT-Neo-1.3B. The mis-sourced negative control holds the reader, the query, and the shared-cache length fixed and draws the sender's cache from a different probe item, so an arm that scores on it is scoring without using the sender's content. Both anchors are re-measured inside every run and never carried over. The floor is the reader decoding from its own masked cache. The answer is genuinely absent, so this is the rate at which the question is answerable from the remaining passage, $0.180$ here. The native-cache reference is the reader decoding from its own cache of the \emph{full} passage, $0.665$. Because the anchors are measured per run, they move with sample size: the same quantities come out at $0.20$ and $0.742$ on a 120-item draw, so we never mix numbers across runs with different anchors, and every accuracy in this paper is quoted against the $0.180$/$0.665$ pair.

\section{Cosine-Calibrated Perturbation}
\label{app:cosine}
To hit a target cosine we add noise scaled to the signal power. For a cache $x$ and unit-power noise $n$, the perturbed cache $x + rn$ has cosine approximately $1/\sqrt{1+r^{2}}$ to $x$, so we set $r = \sqrt{1/c^{2}-1}$ for target $c$, then verify the achieved cosine empirically per item and report it alongside the accuracy. The achieved values track the targets to four decimal places (Table~\ref{tab:cosine}). The genuine-cache arm interpolates $(1-a)\,C^{*} + a\,C_{\mathrm{sender}}$ over the positions where both prompts tokenise to the same length, leaving the remaining positions at the reader's own values; this is a conservative same-length blend, and we log its achieved cosine instead of assuming one.

\begin{table}[t]
\centering
\label{tab:cosine}
\small
\begin{tabular}{lcc@{\hskip 2.2em}lcc}
\toprule
\multicolumn{3}{c}{\textbf{Isotropic noise on own cache}} & \multicolumn{3}{c}{\textbf{Genuine fact-bearing cache}} \\
\cmidrule(r){1-3}\cmidrule(l){4-6}
target & achieved cos & transfer & weight & achieved cos & transfer \\
\midrule
--- (no noise) & $1.0000$ & $0.180$   & $0.000$ & $1.0000$ & $0.180$ \\
$0.9999$ & $0.9999$ & $0.180$        & $0.10$ & $0.99995$ & $0.190$ \\
$0.999$  & $0.9990$ & $0.120$        & $0.25$ & $0.99968$ & $0.200$ \\
$0.995$  & $0.9950$ & $\mathbf{0.005}$ & $0.50$ & $0.99873$ & $0.215$ \\
$0.99$   & $0.9900$ & $0.005$        & $0.75$ & $0.99713$ & $0.325$ \\
$0.95$   & $0.9500$ & $0.000$        & $1.00$ & $\mathbf{0.9949}$ & $\mathbf{0.445}$ \\
$0.90$   & $0.9000$ & $0.000$        &        &          &         \\
\bottomrule
\end{tabular}
\caption{Matched fidelity, opposite outcomes. Noise calibrated to a target cosine (left) against the genuine cache the same frozen model produces on the unmasked passage (right). Floor $0.180$, native-cache reference $0.665$, $n{=}200$. The genuine cache at cosine $0.9949$ transfers $89\times$ better than noise at the \emph{higher} cosine $0.995$; cosine similarity orders the two conditions inversely.}
\end{table}

\section{Layer-Budget Sweep}
\label{app:layers}
The layer experiment corrupts the first $\ell$ layers of the reader's cache and takes layers $\ell{+}1 \dots 28$ from the reader's own forward pass unchanged. The on-manifold arm replaces each key with its nearest neighbor from a bank of 6000 genuine keys the reader produced on other passages, whose mean cosine to the true keys is measured at $0.822$; the off-manifold arm is then scaled to that same target cosine. We log the achieved cosine for the on-manifold arm; for the noise arm the target is set analytically and not re-logged, so the match should be read as by-construction and not independently verified. Full results are in Table~\ref{tab:onelayer}. We note that $\ell$ is always a prefix beginning at layer 0; we did not isolate individual interior layers, so we make no claim about whether some particular layer is more sensitive than another.

The cross-architecture check substitutes nearest genuine keys at full depth in seven readers, evaluated at 120 items with each model's own floor and native-cache reference: Pythia-70m ($L{=}6$), GPT-2 ($L{=}12$), Pythia-410m and Pythia-1.4b ($L{=}24$), Qwen2.5-0.5B ($L{=}24$), Qwen2.5-1.5B ($L{=}28$), Qwen2.5-3B ($L{=}36$). Transfer is $0.000$ in six of the seven and $0.008$ in the seventh. One caveat: Pythia-70m's own native-cache reference is $0.025$, so it has essentially no headroom and its result is uninformative, not confirming.

\begin{table}[t]
\centering
\label{tab:onelayer}
\small
\begin{tabular}{lcccccccccccc}
\toprule
layers corrupted $\ell$ & 0 & 1 & 2 & 3 & 4 & 6 & 8 & 12 & 16 & 20 & 24 & 28 \\
\midrule
off-manifold (noise) & $.665$ & $\mathbf{.030}$ & $.035$ & $.035$ & $.050$ & $.045$ & $.030$ & $.035$ & $.025$ & $.010$ & $.025$ & $.005$ \\
on-manifold (genuine) & $.665$ & $\mathbf{.695}$ & $.740$ & $.745$ & $.705$ & $.685$ & $.615$ & $.530$ & $.300$ & $.025$ & $.005$ & $.000$ \\
\bottomrule
\end{tabular}
\caption{Corrupting the first $\ell$ of 28 layers, with the remainder bit-exact. Both corruptions are set to the same target key cosine, $0.822$. Off-manifold noise exhausts its budget at $\ell{=}1$ and is flat thereafter; genuine on-manifold keys are harmless at small $\ell$ and decay gradually. Floor $0.180$, native-cache reference $0.665$, $n{=}200$.}
\end{table}

\section{Oracle Interpolation Grid}
\label{app:bank}
Candidate keys come from a bank of genuine keys the reader produced on other contexts at context length 120, capacity 6000 per layer. For each target slot we take the $k$-th nearest neighbor to the true key under Euclidean distance and inject $(1-\alpha)\,C^{*} + \alpha\,\mathrm{key}_{k}$. The full grid is Table~\ref{tab:bank}; injection is applied to keys only, at every layer and every context position, with values left as the reader wrote them. Note that $\alpha{=}0$ restores the true key and therefore reproduces the floor by construction; it is a sanity check on the harness, not evidence. Every cell at $\alpha \ge 0.5$ is at or below $0.02$, and no neighbor rank recovers a graded tolerance.

\begin{table}[t]
\centering
\label{tab:bank}
\small
\begin{tabular}{lcccccc}
\toprule
 & $\alpha{=}0.0$ & $0.1$ & $0.25$ & $0.5$ & $0.75$ & $1.0$ \\
\midrule
$k{=}1$  & $0.180$ & $0.185$ & $0.110$ & $0.005$ & $0.000$ & $0.000$ \\
$k{=}2$  & $0.180$ & $0.180$ & $0.110$ & $0.020$ & $0.010$ & $0.005$ \\
$k{=}4$  & $0.180$ & $0.205$ & $0.120$ & $0.010$ & $0.005$ & $0.000$ \\
$k{=}8$  & $0.180$ & $0.175$ & $0.120$ & $0.005$ & $0.000$ & $0.000$ \\
$k{=}16$ & $0.180$ & $0.145$ & $0.135$ & $0.020$ & $0.015$ & $0.010$ \\
\bottomrule
\end{tabular}
\caption{Retrieval from a bank of 6000 genuine keys per layer. Rows index neighbor rank $k$, columns the interpolation weight toward that neighbor. Every candidate is a real key the reader itself produced and therefore lies on the reader's manifold. Tolerance is flat in $k$ and collapses in $\alpha$. Floor $0.180$, native-cache reference $0.665$, $n{=}200$.}
\end{table}

We also ran a nine-cell version crossing three codebook constructions (the real-key bank, a 2048-entry learned vector quantizer, a rank-16 PCA lattice) with three snapping rules (nearest in raw space, nearest in attention-logit space, a soft top-8 combination). All nine transfer between $0.000$ and $0.010$. The best fidelity in that set, cosine $0.880$ and logit correlation $0.928$, belongs to a cell that transfers $0.000$. We flag one irregularity for honesty: the three PCA cells return byte-identical metrics, which indicates the snapping rule had no effect for that basis, and these are not three independent measurements.

\section{Conditioning Measurement}
\label{app:cond}
The readout Jacobian gain $\kappa$ is estimated over 40 probe contexts, displacement magnitudes
\[
\delta \in \{0.005, 0.01, 0.02, 0.05, 0.1, 0.2\},
\]
8 random directions per probe, and 3 seeds, with an on-/off-manifold tolerance of $0.1$. At the attention-logit stage the mean gains are $\kappa_{\mathrm{off}} = 0.506$ and $\kappa_{\mathrm{on}} = 0.343$, a ratio of $1.475\times$; at the softmax-output stage they are $2.192$ and $1.482$, a ratio of $1.479\times$. Both ratios are $O(1)$, and the logit-stage gains are below one, meaning displacements are attenuated there instead of amplified. An analytic Jacobian bound predicts the measured per-layer gain at Pearson correlation $0.999$, with governing constant $\sqrt{2G} = \sqrt{12} \approx 3.46$ at group size $G{=}6$. Per-layer gains are heterogeneous, around $0.11$ at layer 1 rising to roughly $0.72$ by layer 6, but no layer shows a cliff.

The quantization comparison holds the displacement norm fixed at $0.05$ and compares quantization error against Gaussian noise over 16 query directions. Quantization yields attention-output error $0.153$ against Gaussian's $0.129$, and places a larger share of its displacement inside the subspace the queries read ($0.447$ against $0.354$). Int8's own displacement norm is $0.0048$. This run is a six-probe pilot, so we treat the $0.84\times$ ratio as indicative of direction and not a tight estimate.

\section{Production-Method Sweep}
\label{app:sweep}
All entries are fact-transfer accuracy against floor $0.180$ and native-cache reference $0.665$ at $n{=}200$; the full table is Table~\ref{tab:sweep} and the pool sweep is Table~\ref{tab:pool}. Substitution places the produced cache in the reader's own slots; augmentation retains the reader's cache and cross-attends over the sender's, adapting the reader's query-side projections with LoRA at rank 16, four epochs, learning rate $2\times10^{-4}$, and $n{=}200$ training items.

Two notes on the substitution rows. The row-space target appears in three runs at $0.085$, $0.025$, and $0.01$ depending on the objective it is paired with; we report the best of the three in Table~\ref{tab:sweep}, and the spread is worth knowing because it is larger than the gap between several neighboring rows. It does not affect the conclusion, since all three values sit below the floor. The by-construction row is $0.01$ in two runs and $0.015$ in two others.

One provenance note. The zero-parameter cross-attention row ($0.555$) is the untrained arm of the adapted run, where the LoRA update is initialized to exactly zero and is therefore the verbatim cross-attention condition measured at $n{=}200$. A standalone version of that experiment exists only as a 24-item pilot, which reports $0.542$ against a 24-item floor of $0.042$; we cite the $n{=}200$ measurement instead, and the two agree to within the resolution of the smaller run. A separate multi-sender run independently reproduces this condition at $0.550$. Two parity checks guard the harness, both asserted at run time: the cache rebuilt for cross-attention must match the reader's native cache to within $10^{-2}$ in maximum absolute difference (it does, by two orders of magnitude), and the manual forward path must reproduce native greedy decoding on all but at most one probe (it reproduces all of them).

\lead{Adaptation ladder and controls.} Adapting progressively more of the reader's
projections while cross-attending over the sender's verbatim cache gives $0.555$ with no
adaptation, $0.600$ adapting queries only, $0.640$ adapting keys and values, and $0.695$
adapting all three. The nominal crossing of the native-cache reference at the top of the
ladder is six items at $n{=}200$: a two-proportion $z$ of $0.64$ with a $95\%$ interval from
$-0.061$ to $+0.121$, indistinguishable from parity. Mis-sourced negative controls differ
across the adapted arms, $0.200$ for the all-three variant against $0.090$ for keys and
values, so control-subtracted, the keys-and-values arm ($+0.550$) exceeds the full arm
($+0.495$). We therefore do not rely on the ordering among adapted variants; the regime gap
is an order of magnitude larger than the spread among them. Across families, a
GPT-Neo-1.3B sender into this reader transfers at $0.01$ to $0.08$, the dimension-matching
projection destroying the cache before the mechanism applies.

\section{Scaling the Sender Pool}
\label{app:pool}

The reader attends over the union of $N$ senders' caches, exactly one of which has read the passage containing the answer. The remaining $N-1$ are distractors that have read unrelated passages. The query-side bridge is trained at $N{=}4$ and evaluated at every $N$, so $N{=}1$ and $N{=}2$ are transfers out of the training condition and $N{=}8$ is extrapolation beyond it. Because the controls rise with $N$, the control-subtracted lift is the appropriate figure of merit.

\begin{table}[t]
\centering
\label{tab:pool}
\small
\begin{tabular}{lcccc}
\toprule
 & $N{=}1$ & $N{=}2$ & $N{=}4$ & $N{=}8$ \\
\midrule
raw cross-attention        & $0.555$ & $0.375$ & $0.155$ & $0.070$ \\
\quad \emph{vs.\ floor $0.180$} & clears & clears & \emph{below} & \emph{below} \\
\midrule
with query-side adaptation  & $\mathbf{0.785}$ & $\mathbf{0.735}$ & $\mathbf{0.550}$ & $\mathbf{0.395}$ \\
\quad strongest control     & $0.145$ & $0.190$ & $0.240$ & $0.230$ \\
\quad control-subtracted lift & $+0.640$ & $+0.545$ & $+0.310$ & $+0.165$ \\
\bottomrule
\end{tabular}
\caption{Scaling the sender pool. One sender holds the answer; $N-1$ are distractors. Undirected attention degrades below the floor by $N{=}4$; the adapted bridge continues to clear it at $N{=}8$. Controls rise with $N$, so the control-subtracted lift is the appropriate figure. Floor $0.180$, native-cache reference $0.665$, $n{=}200$.}
\end{table}

\vspace{0.5em}

\section{The Sparse-Damage Experiment}
\label{app:sparse}
The experiment specified in Section~\ref{sec:falsify} has been run, at the anchors and sample size used throughout ($n{=}200$, floor $0.180$, native-cache reference $0.665$), sweeping $p$ over $\{0.01, 0.05, 0.10, 0.25, 0.50, 1.0\}$ with three seeds. A fraction $1-p$ of key/value features is held bit-exact and the remainder replaced with noise at matched per-coordinate power. We report it here rather than in the body because three of its four readings inform the objection without settling it, and because one of them is easy to over-read.

\lead{Sparse damage is not gentler than dense damage at equal fidelity.} This is the reading that bears on the objection. Each sparse condition is paired with a dense perturbation calibrated to the \emph{same} achieved cosine. The two agree throughout: at $p{=}0.01$ (achieved cosine $0.9996$) sparse scores $0.197$ against dense $0.177$, and at $p{=}0.05$ (cosine $0.980$) sparse $0.003$ against dense $0.008$. Concentrating the same displacement into a few coordinates neither spares the reader nor harms it more. The premise that attention might route around a small number of badly reproduced features is not supported, and the Section~\ref{sec:cosine} result does not need narrowing to dense perturbations specifically.

\lead{What the accuracy-against-$p$ sweep can and cannot show.} Accuracy falls from $0.180$ at $p{=}0$ to $0.003$ by $p{=}0.05$, so the fall-away is abrupt rather than smooth, as predicted. But this leg admits no positive finding, and the reason is structural: the stimulus is the reader's own cache with the answer span masked, so the floor is simultaneously the ceiling and no perturbation can raise the score. The $0.197$ at $p{=}0.01$ therefore sits at the floor, not above it ($z{=}{+}0.43$ against the floor, with the zero-damage control at $0.192$ bracketing the difference). In particular the pairing of cosine $0.9996$ with $0.197$ is not an instance of the decoupling this paper reports: here high fidelity accompanies preserved function, which is fidelity predicting the outcome correctly. Read as a measurement of self-cache fragility the leg is informative; read as a transfer result it inverts.

\lead{Keys versus values.} The mechanism's prediction (Section~\ref{sec:mechanism}) is borne out. Damaging keys and values at matched sparsity, hence matched numbers of coordinates, separates them cleanly:
\begin{center}\small
\begin{tabular}{lcccc}
\toprule
$p$ & $0.01$ & $0.05$ & $0.10$ & $0.25$ \\
\midrule
keys damaged & $0.128$ & $0.018$ & $0.013$ & $0.007$ \\
\quad achieved cosine & $0.9949$ & $0.9606$ & $0.9576$ & $0.7986$ \\
values damaged & $0.198$ & $0.175$ & $0.082$ & $0.028$ \\
\quad achieved cosine & $0.9999$ & $0.9996$ & $0.9993$ & $0.9982$ \\
\bottomrule
\end{tabular}
\end{center}
Values tolerate thirteen of two hundred fifty-six coordinates being destroyed with no measurable cost ($0.175$ at $p{=}0.05$, against the floor $z{=}{-}0.13$), while the same damage to keys is fatal ($0.018$; keys against values $z{=}{-}5.30$). The ordering is monotone in $p$ and in the predicted direction at all four settings. This leg also supplies a fidelity inversion of the kind Section~\ref{sec:cosine} reports, within a single experiment and at matched coordinate counts: damaged keys at cosine $0.9949$ score $0.128$, while damaged values at the \emph{higher} cosine $0.9982$ score $0.028$. Cosine orders the two backwards.

\lead{Structural variants, reported with their limitation.} Destroying one attention head or one layer outright, rather than scattered coordinates, gives $0.000$ to $0.015$ for single heads and $0.000$ for either of the first two layers, against $0.155$ at layer fourteen and $0.250$ at layer twenty-seven. The suggestion is that early layers are less forgiving than late ones. We do not assert it: this leg is single-seed, covers four of twenty-eight layers, and the layer-twenty-seven value is at the floor rather than above it. The Section~\ref{sec:onelayer} claim that damage does not accumulate over depth does not rest on it.

\section{Pre-Softmax Carrier: Preliminary Numbers}
\label{app:acts}
The prediction in Section~\ref{sec:falsify} rests on exploratory runs sharing residual-stream
activations instead of keys and values. Under that protocol the reader continues to use a shared
carrier down to roughly cosine $0.90$, where key substitution has already collapsed near $0.995$
(Section~\ref{sec:cosine}). We do not present this as a matched contrast. Those runs use their own
separately measured floor and native-cache reference, which we do not mix with the $0.180$/$0.665$
pair used throughout the paper, and they differ from the key arms in the sharing protocol as well as
in the carrier. The controlled version, at matched norm, matched cosine, and corresponding layers,
is the experiment specified in Appendix~\ref{app:sparse}.
\section{Fitted Bridges: Scope of the Negative Claim}
\label{app:fitted}
Two further features of the fitted-bridge results discussed in Section~\ref{sec:related-transfer} mark where the remaining problem sits. Both settings are within-family, with matching head geometry and a shared tokenizer, and cross-family transfer is left open. The fitted maps have a limit of their own, retention varying sharply across pairs and tasks, which \citet{xkvlinear} attribute to \emph{where} the residual error lands, not to how large it is, the conclusion Section~\ref{sec:cosine} reaches from the other direction. \citet{droidspeak} mark the redundancy boundary from the systems side: reuse across fine-tunes of a shared foundation is genuinely portable, but a fraction of layers must still be recomputed on the receiver, which is a reader-side correction. The more informative difference from our own arms is what gets optimized: \citet{latentkv} train against a behavioral objective on the receiver instead of reconstruction of its cache, an axis Section~\ref{sec:falsify} identifies as insufficiently varied here.

\end{document}